\documentclass[letterpaper, 10 pt, conference]{ieeeconf}  
\usepackage[draft]{pdfpages}

\IEEEoverridecommandlockouts                              

\usepackage{graphicx} 
\usepackage{amsmath} 
\usepackage{amssymb}  
\usepackage[ruled]{algorithm2e} 
\usepackage{subcaption} 
\usepackage{color} 
\usepackage[noadjust]{cite} 
\usepackage{hyperref}

\usepackage{array }
\usepackage{booktabs} 
\graphicspath{{figures/}} 

\DeclareCaptionLabelSeparator{periodspace}{.\quad} 
\title{\LARGE \bf
Directional grid maps: modeling multimodal angular uncertainty\\ in dynamic environments
}

\author{Ransalu Senanayake$^{1}$ and Fabio Ramos$^{1}$ \\
\small $^1$The University of Sydney, Australia
\thanks{$^1$Email: {\tt\small rsen4557@uni.sydney.edu.au} }
}

\begin{document}

\maketitle
\thispagestyle{empty}
\pagestyle{empty}

\setlength{\abovedisplayskip}{3pt}
\setlength{\belowdisplayskip}{3pt}
\setlength{\textfloatsep}{10pt plus 1.0pt minus 2.0pt}

\begin{abstract}

Robots often have to deal with the challenges of operating in dynamic and sometimes unpredictable environments. Although an occupancy map of the environment is sufficient for navigation of a mobile robot or manipulation tasks with a robotic arm in static environments, robots operating in dynamic environments demand richer information to improve robustness, efficiency, and safety. For instance, in path planning, it is important to know the direction of motion of dynamic objects at various locations of the environment for safer navigation or human-robot interaction. In this paper, we introduce directional statistics into robotic mapping to model circular data. Primarily, in collateral to occupancy grid maps, we propose \emph{directional grid maps} to represent the location-wide long-term angular motion of the environment. Being highly representative, this defines a probability measure-field over the longitude-latitude space rather than a scalar-field or a vector-field. Withal, we further demonstrate how the same theory can be used to model angular variations in the spatial domain, temporal domain, and spatiotemporal domain. We carried out a series of experiments to validate the proposed models using a variety of robots having different sensors such as RGB cameras and LiDARs on simulated and real-world settings in both indoor and outdoor environments. 

\end{abstract}

\section{INTRODUCTION}
\label{sec:intro}

Safe operation of robots in dynamic environments where humans, vehicles, and other robots operate is central to full autonomy. Spatial information alone is not sufficient in complex environments. This is because the prediction of future events drives decision making whilst properly managing the risk of collisions. Although conventional mapping techniques represent the space in terms of the probability of occupancy \cite{Elfes87,Ramos15}, they do not explicitly capture the patterns in the motion direction of dynamic objects such as people, cars, and cyclists. Understanding and modeling directions are complicated and cannot be treated with conventional techniques from linear statistics as the treatment of angular quantities requires that distributions be mapped into hyperspheres, in a set of techniques known as directional statistics \cite{DirectStat}.       

\begin{figure}[t]
  \centering
  \includegraphics[width=\linewidth]{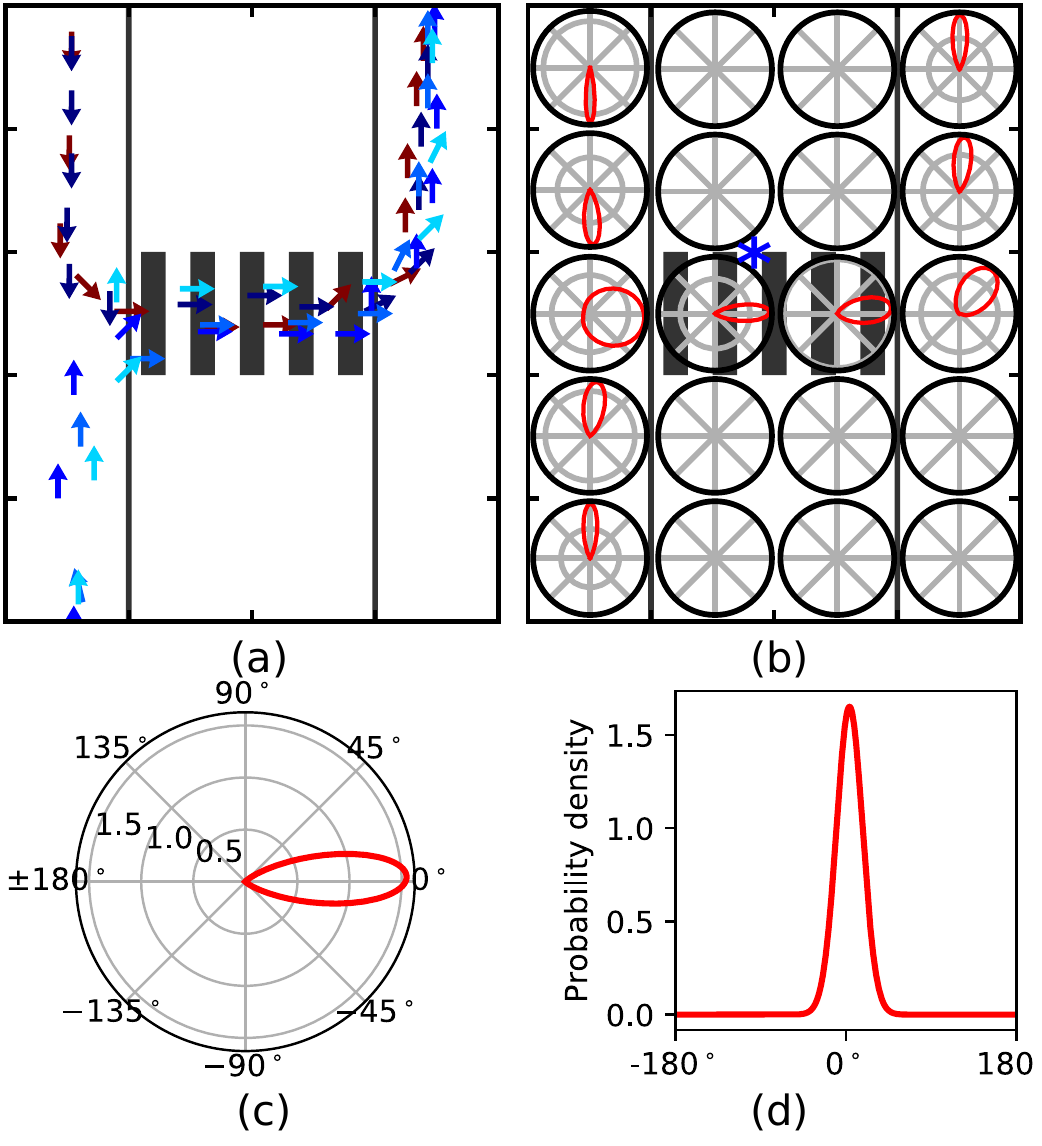}
  \caption{Motivation for directional mapping. (a) Plan view of a road where simulated human coming from two directions walk across a crosswalk and head towards a different direction. We are interested in modeling directions people move after taking observations over a period of time (b) The world is divided into a $5\times4$ grid. The \emph{directional distribution}---a probability distribution over directions $[-180^\circ,180^\circ]$---at different cells modeled using the proposed method (DGM) is illustrated using polar plots. For any location/cell in the space, such a directional distribution exists. The correspondence between the direction of arrows and the direction of polar plots can be observed. (c) To elaborate, the polar plot marked in $*$ on the grid map is emphasized. This shows the probability density for different angles at the particular cell---the probability density increases as go farther away from the center. (d) The equivalent unwrapped probability function with support $[-180^\circ,180^\circ]$ is given for clarity. }
  \label{fig:moti}
\end{figure}

In general, a robot requires a map for path planning and safe navigation. To this end, the most common approach is to represent the actual geometry of the environment as floor plans \cite{Elfes87} or 3D models \cite{henry2012rgb, bohg2011multi} in the metric space, though graph-based approaches also exist \cite{yin2014graph}. These metric maps are typically built using data collected from RGB cameras or depth sensors such as LiDAR  \cite{Elfes87,hornung2013octomap}. 

The basic information about the environment a robot requires to maneuver is to know which areas of the environment are occupied and which areas are not. To model this, in his seminal paper, Elfes \cite{Elfes87} proposed occupancy grid maps---the environment is divided into a grid and occupancy probability of each cell is updated as beam reflections are collected from a depth sensor such as sonar or LiDAR. Later, continuous scalar-field representations have been proposed \cite{GPOMIJRR,Ramos15}. However, all of these methods assumed a static environment where the only moving object in the environment is the robot. To build a static occupancy map in the presence of a few dynamic objects, \cite{hahnel2003,burgard2007mobile} proposed to filter dynamic objects as a preprocessing step and then map the occupancy.

More recently, rather than considering dynamic objects as nuisances, they have been incorporated into the map in order to model the long-term occupancy \cite{walcott2012dynamic,rans_2016,rans2017icra,senanayake2017bayesian} and understand occupancy patterns \cite{saarinen2012,meyer2012,wang2015modeling,krajnik2014}. Nevertheless, unlike in static environments, occupancy is not the only information that can be extracted in dynamic environments. Supplementing additional information about the dynamics of the environment could hugely benefit path planning and object detection algorithms \cite{walcott2012dynamic,bore2017object}. For this purpose, information rich maps can be developed by modeling the uncertainty of directions, speed, texture, etc. of all locations of the environment in addition to occupancy information. Consider an instance as in Fig.~\ref{fig:moti} (a) where simulated humans walking in roadsides and a crosswalk. If the robot knows about the angles people turn, path planning algorithms can be designed to plan ahead and to make efficient and safer maneuvers. As shown in Fig.~\ref{fig:moti}b, in this paper, we propose a novel technique to model directional uncertainty of the environment at different locations observed over time. 

\cite{human_simon} proposed to model human walking paths by introducing a Gaussian process prior over directions and thereby implicitly constructing a field representation of angular movements. This formulation has three main limitations: 1) because the angle is assumed to be $(-\infty,+\infty)$, predicted angles can be totally invalid, 2) it is assumed that movements at a given location occur in only one direction which is not practical for robotics applications as the robot, human, or vehicles in the environment could move in any direction, and 3) being a Bayesian nonparametric model, the algorithm becomes slower as more data are captured. On the other hand, the objective of all these approaches is to make short-term future predictions such as tracking rather than building information-rich long-term maps that can be used for path planning or navigation.

In our approach, highlighting the importance of dispersion of data, the directions are represented by a probability distribution that, 1) has a valid support of $[-\pi,\pi]$ and 2) can model multi-directional movements. Having a finite number of distributions laid over the longitude-latitude space using a grid, it is possible to infer the probability of motion for any direction for each such location. This directional information can be plausibly used to extract paths as well as variously regulate path planners to avoid high-risk areas or to follow the direction of traffic. Although incorporation of such probability distribution into control algorithms is beyond the focus of this paper, recent techniques have shown how to embed probability distributions to improve path planning and navigation \cite{marinhofunctional,dong2016motion,norouzi2016probabilistic,mukadam2017simultaneous}. Further, incorporating such prior information is the key in Bayesian statistical methods and prior information can be effectively used in online learning in robotics \cite{senanayake2017bayesian,bui2017streaming}. Additionally, such probabilistic models naturally account for noises and imperfections in sensors and pre-processing algorithms.

Despite the importance of modeling the stochasticity of angles in robotics, it has hardly been discussed previously. Therefore, introducing directional statistics into robotics to model angular data, we present a statistical method:
\begin{enumerate}
\item to model multi-modal directional uncertainty in dynamic environments without obtaining spurious outputs as in current methods \cite{human_simon};
\item to quantitatively analyze spatial variations, temporal variations, and spatiotemporal variations; 
\item that does not require heuristic parameter tuning i.e. ready for real-world usage without any significant modifications
\end{enumerate}

Having discussed the motivation for our work in Section~\ref{sec:intro}, Directional Statistics are introduced in Section~\ref{sec:dir} as preliminaries for the following sections. Data preprocessing steps required for mapping is detailed in Section~\ref{subsec:pre}. Then, the basic method is introduced in Section~\ref{subsec:uni} assuming that all movements are almost uni-directional such as one-way roads. Next, in Section~\ref{subsec:multi}, the theory is generalized to model multi-directional movements i.e. when there are no definite paths or dynamic objects can move in arbitrary directions such as in indoor environments or sidewalks. Experimental results are reported in Section~\ref{sec:exp} followed by discussions and conclusions in Sections~\ref{sec:dis} and ~\ref{sec:conc}, respectively.

\section{DIRECTIONAL STATISTICS}
\label{sec:dir}

\begin{figure}[b]
  \centering
  \includegraphics[width=\linewidth]{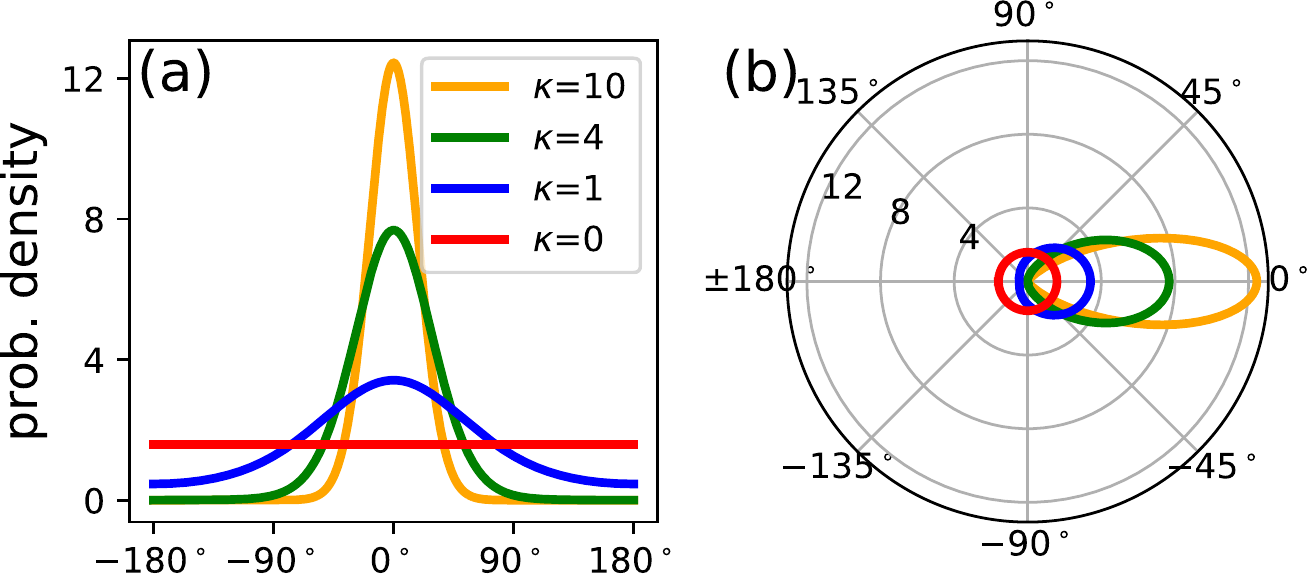}
  \caption{The probability density functions of von Mises directional distribution (a) The effect of the concentration parameter $\kappa$ for a fixed mean $\mu=0^\circ$. The support of von Mises distribution is $[-\pi,\pi]$. For instance, when estimating the direction of a moving object from noisy sensor measurements, a conventional Gaussian distribution which has a support $(-\infty,\infty)$ cannot be used because the actual range of directions is $[-\pi, \pi]$. (b) Corresponding polar plots.}
  \label{fig:kappa}
\end{figure}

In this section, we introduce directional statistics where observations lie on a circle of unit radius, or in high dimensional scenarios, on a hypersphere of unit vector in the plane \cite{DirectStat}. In order to deal with circular data, directional statistics was initially developed in physics and astronomy \cite{vonMis1918} \cite{MardiaEdw1982}, and have successful applications in meteorology \cite{mete}, biostatistics \cite{van2013vector} etc. 

Although there are several approaches to model directional data \cite{gilitschenski2016unscented}, we opted for the von Mises distribution \cite{vonMis1918} because, i) it has all advantages of the exponential family of distributions, ii) analogous to a Gaussian distribution with more intuitive parameters, and iii) \emph{sufficient statistics} can be obtained explicitly \cite{DirectStat}. These properties will be intermittently discussed in Sections~\ref{sec:dgm} and ~\ref{sec:dis}. The probability density function of the von Mises distribution is given by (\ref{eq:vonM}),
\begin{equation}
  \mathcal{VM}(\theta; \mu, \kappa) := \frac{1}{2\pi J_0(\kappa)} \exp{\big( \kappa \cos(\theta - \mu) \big) },
 \label{eq:vonM}
\end{equation}
\noindent where $\mu$ is the mean direction parameter (analogous to mean in a Gaussian distribution) and $\kappa$ is the concentration parameter (weakly analogous to the reciprocal of variance in a Gaussian distribution). $J_0(\kappa)$ is the $0^\text{th}$ order and $1^\text{st}$ kind modified Bessel function given by (\ref{eq:Bas}),

\begin{equation}
J_0(\kappa) := \sum_{p=0}^\infty \frac{1}{{p!}^2} \Big( \frac{\kappa}{2} \Big)^{2p}.
\label{eq:Bas}
\end{equation}

Together with the cosine term in (\ref{eq:vonM}), the modified Bessel function attenuates the function and keeps the support in $[-\pi, \pi]$. The effect of the $\kappa$ parameter is illustrated in Fig.~\ref{fig:kappa}.

Considering a dataset with $N$ directions $\mathcal{D} = \{ \theta_i\}_{i=1}^N$, let us define the mean directional components,
\begin{equation}
\bar{C} := \sum_{i=1}^N \cos\theta_i \quad \text{ and } \quad \bar{S} := \sum_{i=1}^N \sin\theta_i.
\end{equation}

Then, the \emph{mean direction} is given by,
\begin{equation}
\bar{\theta} = 
\begin{cases}
\arctan (\bar{S}/\bar{C}), & \text{if } \bar{C} \geq 0 \\
\arctan (\bar{S}/\bar{C}) + \pi, & \text{otherwise}
\end{cases}
\end{equation}
and the \emph{mean resultant length} is given by,
\begin{equation}
\bar{R} = \sqrt[]{\bar{C}^2 + \bar{S}^2}.
\end{equation}
Note that $\bar{R} \in [0,1]$ and the more homogeneous the directions are, the higher the $\bar{R}$ is. The \emph{circular variance} is defined as $\bar{V} := 1 - \bar{R}$.

\section{DIRECTIONAL GRID MAPS}
\label{sec:dgm}

Being analogous to occupancy grid maps \cite{Elfes87}, we introduce \emph{directional grid maps} (DGM) in this section. To formally define, a DGM is a multi-dimensional field that maintains probability measures given by a probability density function of the directional uncertainty of the cells in a spatial lattice. With the proposed method, we answer the following questions:
\begin{enumerate}
\item What are the overall directions of motion in different places in the environment when observed over a period of time? i.e. longterm spatiotemporal analysis;
\item What is the overall direction of motion in the entire environment at a specific time? i.e. spatial analysis; 
\item What is the distribution of directions of a moving object? i.e temporal analysis.
\end{enumerate}

\subsection{Pre-processing}
\label{subsec:pre}

The inputs to build an occupancy map are occupied points and the free points in the line of sight of LiDAR \cite{Elfes87,GPOMIJRR}. However, inputs to build a DGM are the angle of motion at longitude-latitude locations at different time steps: $\theta(longitude, latitude, time)$. For simplicity and to be used in collateral to occupancy grid maps, we discretize the world and assign longitude-latitude pairs to the cell they belong to. Therefore, the inputs  are $\theta(cell,time)$. 

Depending on the sensor type, for each time frame, $\theta$ values or the optical flow can be extracted by any of the commonly used existing methods. To name a few, data association followed by direct angle estimation or Gaussian process regression \cite{rans_2016}, tracking algorithms such as Kalman filters, particle filters, mean-shift-tracking, dense optical flow, etc. \cite{farneback2003two,sivaraman2013looking}. Once $\theta(cell,time)$ are extracted, for the computational convenience of answering questions detailed in Section~\ref{sec:dgm}, they are stored with tracker identities, if exists, in a spatiotemporal database \cite{spatiotemporal_db} indexed by space and time keys. 

\subsection{Learning uni-modal movements (DGM-VM)}
\label{subsec:uni}

Without loss of generality, for the sake of simplicity to introduce the method, in this section, we assume the average movements in the environment occur in approximately one direction. The more general case is introduced in Section~\ref{subsec:multi}.

Consider a dataset with $N$ directions $\mathcal{D} = \{ \theta_i\}_{i=1}^N$ Assuming i.i.d. of $\theta$, the log-likelihood of the von Mises distribution introduced in (\ref{eq:vonM}) is given in (\ref{eq:likeli}),
\begin{align}
\mathcal{L}(\mu, \kappa; \mathcal{D}) 
&= \log \bigg( \prod_{i=1}^N \frac{1}{2\pi J_0(\kappa)} \exp{\big( \kappa \cos(\theta_i - \mu) \big) } \bigg) \nonumber\\
&= - N\log2\pi - N\log J_0(\kappa) + \kappa \sum_{i=1}^N \cos(\theta_i - \mu)  \nonumber \\
&= - N\log2\pi - N\log J_0(\kappa) + \kappa N \bar{R} \cos(\bar{\theta} - \mu).  \nonumber \\
\label{eq:likeli}
\end{align}

The objective is to learn $\mu$ and $\kappa$ given $\mathcal{D}$ to maximize the log-likelihood, i.e. maximum likelihood estimate (MLE). Intuitively, for a given dataset, MLE adjusts its parameters $\mu$ and $\kappa$ to set the higher values of the probability density function align with more probable data points. These optimal parameter values can be computed by (\ref{eq:opt}),
\begin{equation}
 (\mu_*, \kappa_*) = \mathrm{argmax}\mathcal{L}(\mu, \kappa; \mathcal{D}) =  \big( \bar{\theta}, A^{-1}(\bar{R}) \big)
 \label{eq:opt}
\end{equation}
where $A(\cdot) = \frac{J_1(\cdot)}{J_0(\cdot)}$. To derive this, take the derivative of $\mathcal{L}$ w.r.t. the parameters and equate to zero, 

\begin{align}
\frac{\partial \mathcal{L}}{\partial \mu} 
&= \kappa N \bar{R} \sin(\bar{\theta} - \mu) = 0 \implies \mu_* = \bar{\theta},
\end{align}

\begin{align}
\frac{\partial \mathcal{L}}{\partial \kappa} 
&= - N \frac{J_0^\prime (\kappa)}{J_0(\kappa)} + N \bar{R} \cos(\bar{\theta} - \mu) \nonumber \\
&= - N \frac{J_1(\kappa)}{J_0(\kappa)} + N \bar{R} \cos(\bar{\theta} - \mu) \nonumber \\
&= - N A(\kappa) + N \bar{R} \cos(\bar{\theta} - \mu). 
\end{align}

Setting $\frac{\partial \mathcal{L}}{\partial \kappa} = 0$ and $\mu_* = \bar{\theta}$ $\implies \kappa_* = A^{-1}(\bar{R})$. To approximate $\kappa_*$,
\begin{equation*}
\kappa_* \approx \begin{cases}
     2\bar{R} + \bar{R}^3 + \frac{5}{6} \bar{R}^, & \text{ for small } \bar{R} \\
    0.5(1-\bar{R})^{-1}, & \text{otherwise}.
\end{cases}
\end{equation*}
For empirical values of "small," refer \cite{DirectStat,MardiaEdw1982}. Alternatively, it is also possible to maximize (\ref{eq:likeli}) w.r.t. the parameters using stochastic gradient descent.

\subsection{Learning multi-modal movements (DGM-VMM)}
\label{subsec:multi}
\begin{figure}[t]
  \centering
  \includegraphics[width=\linewidth]{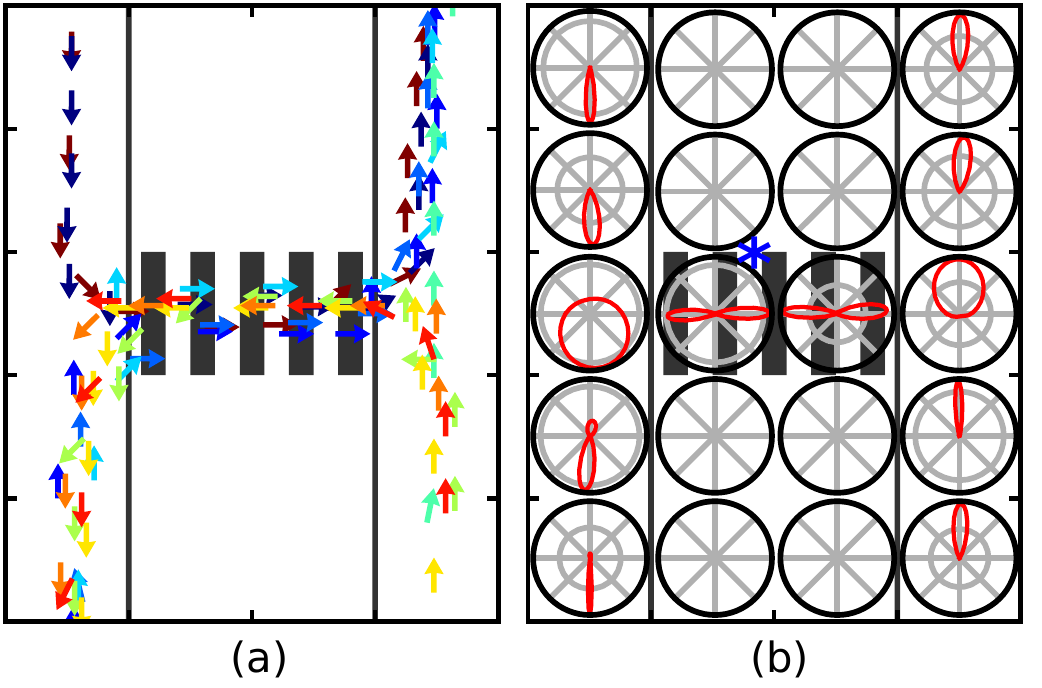}
  \caption{Mapping multi-modality. (a) More human paths are added to Fig.~\ref{fig:moti} so that human walk in different directions in some places e.g. crosswalk (b) DGM modeled using the mixture of von Mises distributions. With comparison to Fig.~\ref{fig:moti} (a) which only has a single von Mises distribution, observe that some cells in (b) have two lobes indicating the method's capacity to learn bimodal movements. The middle-left cell shows a more circular distribution because movements occur in different directions.}
  \label{fig:multi}
\end{figure}

The method in section~\ref{subsec:uni} assumes that movements occur only in one direction. Although this assumption might be applicable for roads with vehicles running in dedicated lanes, such an assumption is not generally suitable for crosswalks, sidewalks, manipulators, or aerial vehicles (Fig.~\ref{fig:multi}). Therefore, in order to capture multi-directional movements, we use the convex combination of a mixture of $M$ directional distributions. The probability density function of such a mixture with $M$ von Mises distributions is given by (\ref{eq:vonMM}).
\begin{equation}
\mathcal{VMM}(\theta; \alpha, \mu, \kappa) := \sum_{m=0}^M \alpha_m  \mathcal{VM}(\theta; \mu_m, \kappa_m),
\label{eq:vonMM}
\end{equation}

\noindent with $\sum_{m=0}^M \alpha_m = 1$ for $\alpha_m \geq 0$ to guarantee $\mathcal{VMM}$ is a valid probability density function.

However, there is no closed-form solution to find the optimal parameter set $\{(\alpha_m, \mu_m, \kappa_m)\}_{m=1}^M$. Therefore, as with Gaussian mixture models, the Expectation-Maximization (EM) algorithm can be used \cite{dhillon2003modeling}. This is an iterative procedure where the posterior is estimated using the parameters estimated in the previous iteration, and the parameters are updated using the estimated posterior in the current iteration. Once the parameters do not significantly change over iterations, the optimization procedure can be stopped. This is detailed in Algorithm~\ref{algo:1}. 

The naive EM algorithm only learns the parameters, not the number of mixture components $M$. Although it is possible to preset it as a fixed number, in order to make the algorithm faster and to make mapping fully autonomous, we made use of DBSCAN \cite{ester1996density} clustering technique and initialized $\{\mu_m\}_{m=1}^M$ with cluster centers, and then optimized using the EM algorithm. Unlike the popular k-means algorithm where the user requires to provide the number of clusters, DBSCAN determines it using the density of data points which is indeed our requirement.

In a similar fashion to a Gaussian mixture model \cite{Bishop2006}, in the E-step, the elements of the responsibility matrix are computed as in (\ref{eq:res}).
\begin{equation}
\gamma_{mn} = \frac{\alpha_m \mathcal{VM}_m(\theta_n)}{\sum_{m'=1}^M \alpha_{m'} \mathcal{VM}_{m'}(\theta_n)},
\label{eq:res}
\end{equation}
where $\mathcal{VM}_m$ indicates a von Mises probability density function of the mixture component $m$. Having obtained $\gamma_{mn}$, the objective of the M-step is to learn the parameters using (\ref{eq:1})-(\ref{eq:2}),

\begin{equation}
\alpha_m = \frac{\sum_{n=1}^N \gamma_{mn} }{N},
\label{eq:1}
\end{equation}
\begin{equation}
\mu_m = \frac{\sum_{n=1}^N \gamma_{mn} \theta_n }{\Vert \sum_{n=1}^N \gamma_{mn} \theta_n \Vert},
\end{equation}
\begin{equation}
\kappa_m = A^{-1}\bigg(\frac{\Vert \sum_{n=1}^N \gamma_{mn} \theta_n \Vert}{\sum_{n=1}^N \gamma_{mn}}\bigg),
\label{eq:2}
\end{equation}

\noindent where $A^{-1}(\cdot)$ is the inverse of Bessel function ratios as described in Section~\ref{subsec:uni}.

\begin{algorithm}[h]
 \KwIn{$\{\theta_n\}_{n=1}^N$ }
 $\{\mu_m^{(0)}\}_{m=1}^?$ = DBSCAN($\theta$)    //get density centers\;
 $M=$size$(\{\mu_m^0\})$    //number of mixture components\;
 Initialize $\alpha_m^{(0)} = 1/M$, for $m=1:M$\;
 Initialize $\kappa_m^{(0)} \gtrsim +0$, for $m=1:M$\;
 Initialize $\epsilon \approx +0$\;
 Initialize $i=-1$  //iterations\;
 \While{$\Vert \mu_m^{(i)} - \mu_m^{(i-1)}\Vert \leq \epsilon$}{
 $i \leftarrow i+1$\;
  //E-step\;
  \For{$n=1 \text{ to } N$}{
    \For{$m=1 \text{ to } M$}{ 
      $\hat{\gamma}_{mn}^{(i)}=$\emph{CalcRes}$\big(\theta_n, \alpha_m^{(i-1)}, \mu_m^{(i-1)}, \kappa_m^{(i-1)}\big)$\;
    }
  }
  //M-step\;
  \For{$m=1 \text{ to } M$}{
     $\big(\alpha_m^{(i)}, \mu_m^{(i)}, \kappa_m^{(i)}\big)=$\emph{UpdateParameters}$\big(\hat{\gamma}_{mn}^{(i)}\big)$\;
  }
 }
  \KwOut{$\alpha_m^{(i)}, \mu_m^{(i)}, \kappa_m^{(i)}$ //$\mathcal{VMM}$ (\ref{eq:vonMM})}
 \caption{EM algorithm for multi-modal learning. \emph{CalcRes()} and \emph{UpdateParameters()} are (\ref{eq:res}) and (\ref{eq:1})-(\ref{eq:2}), respectively.}
 \label{algo:1}
\end{algorithm}

\section{EXPERIMENTS}
\label{sec:exp}

\subsection{Experimental setup and evaluation metrics}

As given in Table~\ref{table:datasets}, we used a variety of datasets from simulated and real-world environments having both LiDAR and cameras, to validate different aspects of the proposed methods and answer questions raised in Section~\ref{sec:dgm}.

\begin{table}[h]
  \centering
  \caption{Description of datasets}
    \begin{tabular}{p{0.17\linewidth}|p{0.7\linewidth}}
    \toprule
      Datasets  & Description \\
      \hline
      Unimodal & Similar to \cite{human_simon}, this simulated dataset represents human walking paths which collectively have a {\bf unimodal} directional pattern i.e. at a given location all human walk approximately in the same direction. Observations are assumed to be taken from the top view. (Fig.~\ref{fig:moti} (a))\\
      Multi-modal & This is the {\bf multi-modal} (to be exact, bi-modal) extension to the above unimodal dataset. (Fig.~\ref{fig:multi} (a))\\
      Edinburgh & The publicly available {\bf Edinburgh} Informatics Forum Pedestrian Database (Aug.24) \cite{majecka2009statistical} is used. The setup is an highly dynamic outdoor environment with RGB cameras setup on top to track \cite{majecka2009statistical} people. (Fig.~\ref{fig:edin} (a)) \\
      Kuka & Here, we use the {\bf Kuka} robot arm in the MORSE simulator \cite{morse_simpar_2012}. The location of the end-effector in the 2D space was tracked. Using only two joints, we manipulated the robot to make planar movements to simulate a repetitive task with normally distributed random perturbations to the goal locations. Because of this perturbations, robot's path is slightly different in each of the 20 iterations which results in observing different angles of the end effector in the same location. (Fig.~\ref{fig:kuka_1} (a)) \\
      Corridor & This tracks five people moving in a {\bf corridor} using a moving robot with a LiDAR \cite{romero2017inlida}. (Fig.~\ref{fig:s3} (a)) \\ 
      Intersection & This is similar to the corridor dataset, however in a four-way {\bf intersection} (Fig.~\ref{fig:s5} (a))\\
            Human & The single trajectory of a walking {\bf human} in a MORSE-simulated office environment is tracked. (Fig.~\ref{fig:seql} (a)) \\
      \bottomrule
    \end{tabular}
  \label{table:datasets}
\end{table}

In order to assess models, we used several metrics. In a $M$-mixture of distributions, the expected negative log-likelihood (ENLL) is calculated as the average negative log-likelihood over all data points\cite{ruppert2011statistics} which indicates the likelihood a given data point sampled from the distribution parameterized by $\{(\mu_{m*},\kappa_{m*})\}_{m=1}^M$. For unimodal settings $M=1$. The smaller the NLL or ENLL, the better the model fit is. 

As another metric, average probability density (APD) is considered. Metrics are calculated with a 10-fold cross-validation procedure. For each fold of test data, the probability density is calculated and averaged. Intuitively, if the model has captured the full long-term distribution, it gives a higher APD score because more points are concentrated around the vicinity if that area. Unfortunately, because of the problem is unsupervised learning and having a mixture of distributions, most of the standard tests that are commonly used in linear statistics and robotics with normality assumptions cannot be used for this setting. 

A notebook computer with an Intel Core-i7 processor and an 8 GM RAM was used for experiments. For all experiments, the tolerance parameter of the VMM algorithm $\epsilon$ and the density parameter of the DBSCAN were set to $10^{-6}$ and $0.5$, respectively. When using the Edinburgh dataset with Gaussian process \cite{human_simon} is used in comparisons, a low-rank approximation \cite{gpy2014} had to be used because the Edinburgh dataset contains 76260 data points which are not feasible to fit using the full Gaussian process model. The python code will be available soon: \href{github.com/RansML}{github.com/RansML}.

\subsection{Experiment 1: Validating the EM algorithm}
\begin{figure}[h]
  \centering
  \includegraphics[width=0.7\linewidth]{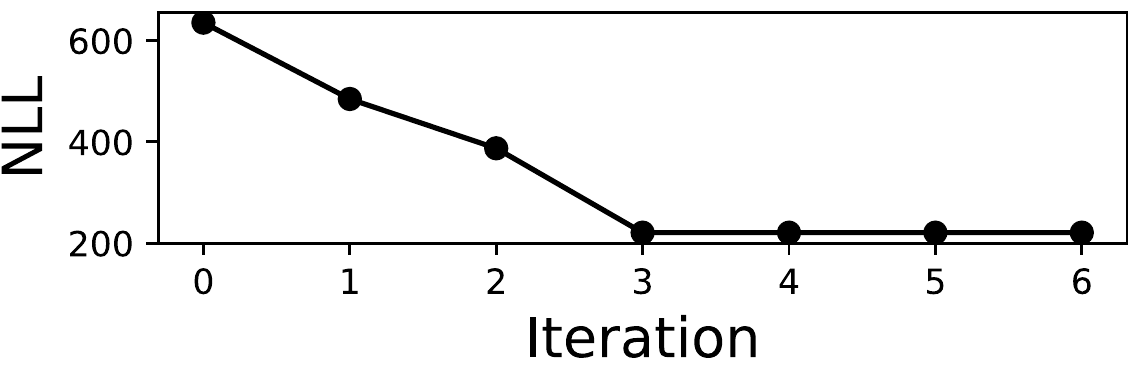}
  \caption{Convergence of the EM algorithm}
  \label{fig:nll}
\end{figure}

Firstly, we show that the iterative EM algorithm described in Section~\ref{subsec:multi} indeed minimizes the NLL over the number of iterations. In Fig.~\ref{fig:nll}, NLL is plotted against the number of iterations for the cell marked with ``$*$" in Fig.~\ref{fig:multi} (b) for the multi-modal dataset. For the Multi-modal dataset, the EM algorithm converged within 5 iterations. The mean squared error (MSE) of angles is reported in Table~\ref{table:comp}. In DGM-VMM, the MSE to the closest mode was considered.

\begin{table}[h]
  \centering
  \caption{Mean squared error (MSE) of angles. }
    \begin{tabular}{c|c|c}
    \toprule
       Method & Unimodal & Multi-modal\\
      \hline
      DGM-VMM & {\bf0.231} & {\bf1.100} \\
      DGM-VM & 0.239 & 2.484 \\
      \bottomrule
    \end{tabular}
  \label{table:comp}
\end{table}

\subsection{Experiment 2: Modeling long-term spatiotemporal effects}

\begin{figure*}[t]
    \centering
    \begin{subfigure}{0.34\linewidth}
    \centering
        \includegraphics[width=\textwidth]{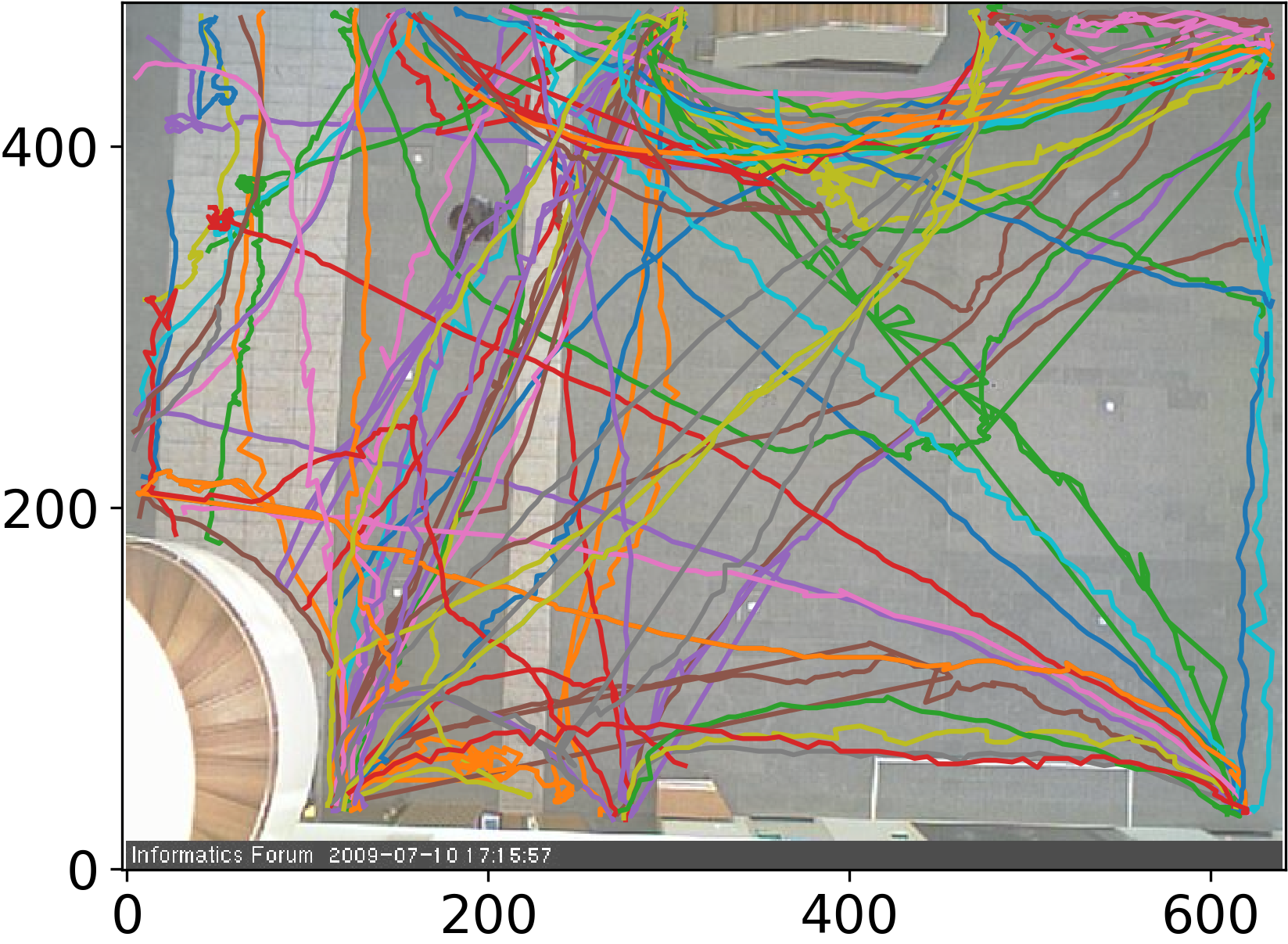}
        \includegraphics[width=0.6\textwidth]{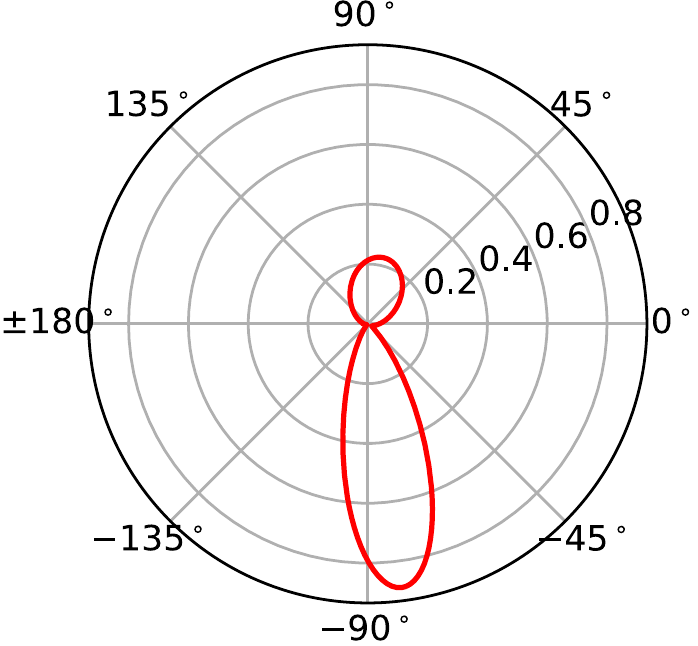}
        \caption{100 randomly selected trajectories overlaid on the real environment. (c) Spatial analysis}
        \label{subfig:edin1}
    \end{subfigure}
    \begin{subfigure}{0.65\linewidth}
    \reflectbox{\rotatebox[origin=c]{180}{
        \includegraphics[width=\textwidth]{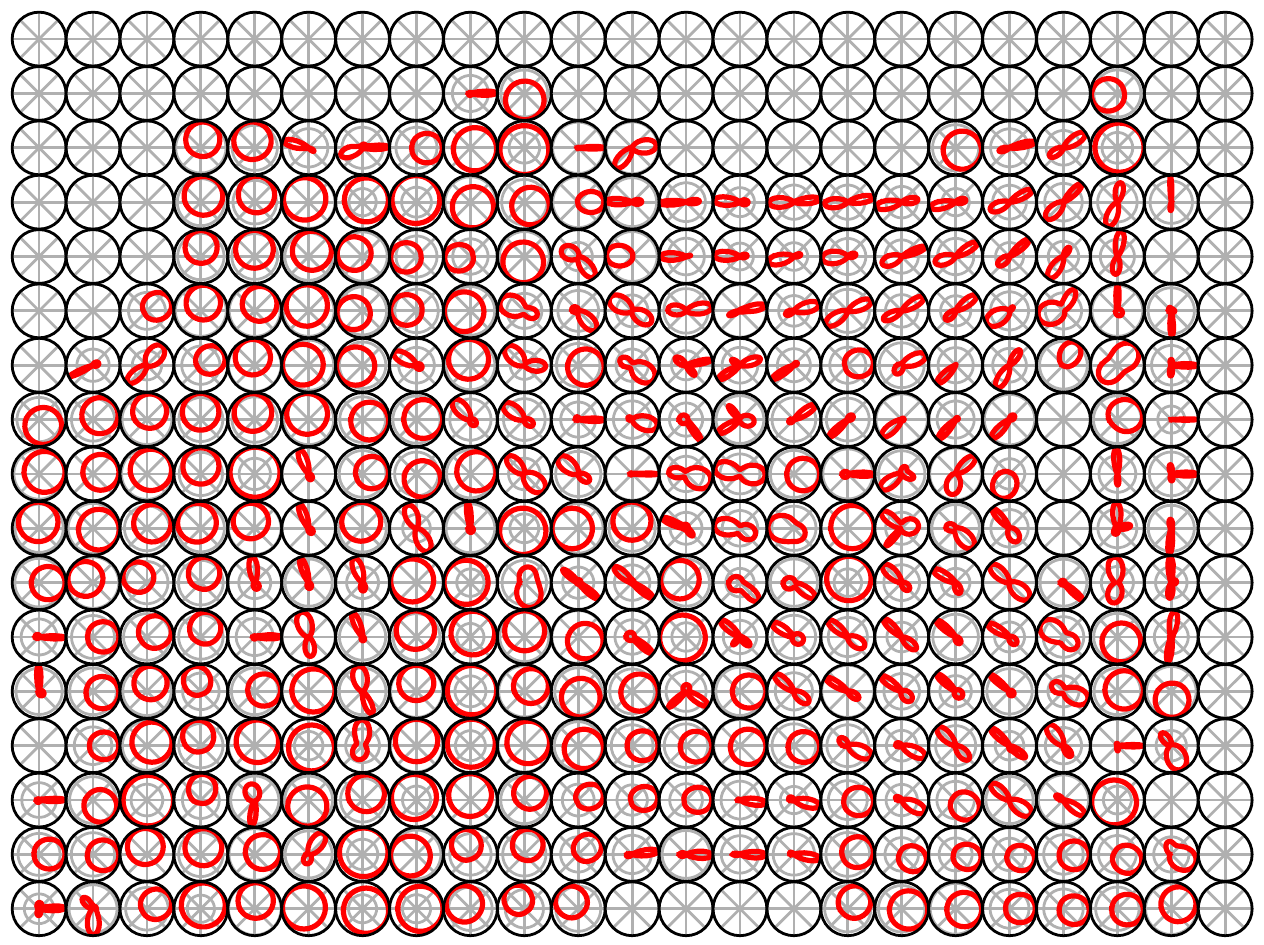} }}
        \caption{VMM-DGM for (a)}
        \label{subfig:edin2}
    \end{subfigure}
    \caption{Edinburgh pedestrian dataset shows how people near the University of Edinburgh's Atrium move from one building to another on a regular day.}
    \label{fig:edin}
    \vspace*{-10pt}
\end{figure*}

\begin{figure}[h]
  \centering
  \includegraphics[width=0.48\linewidth]{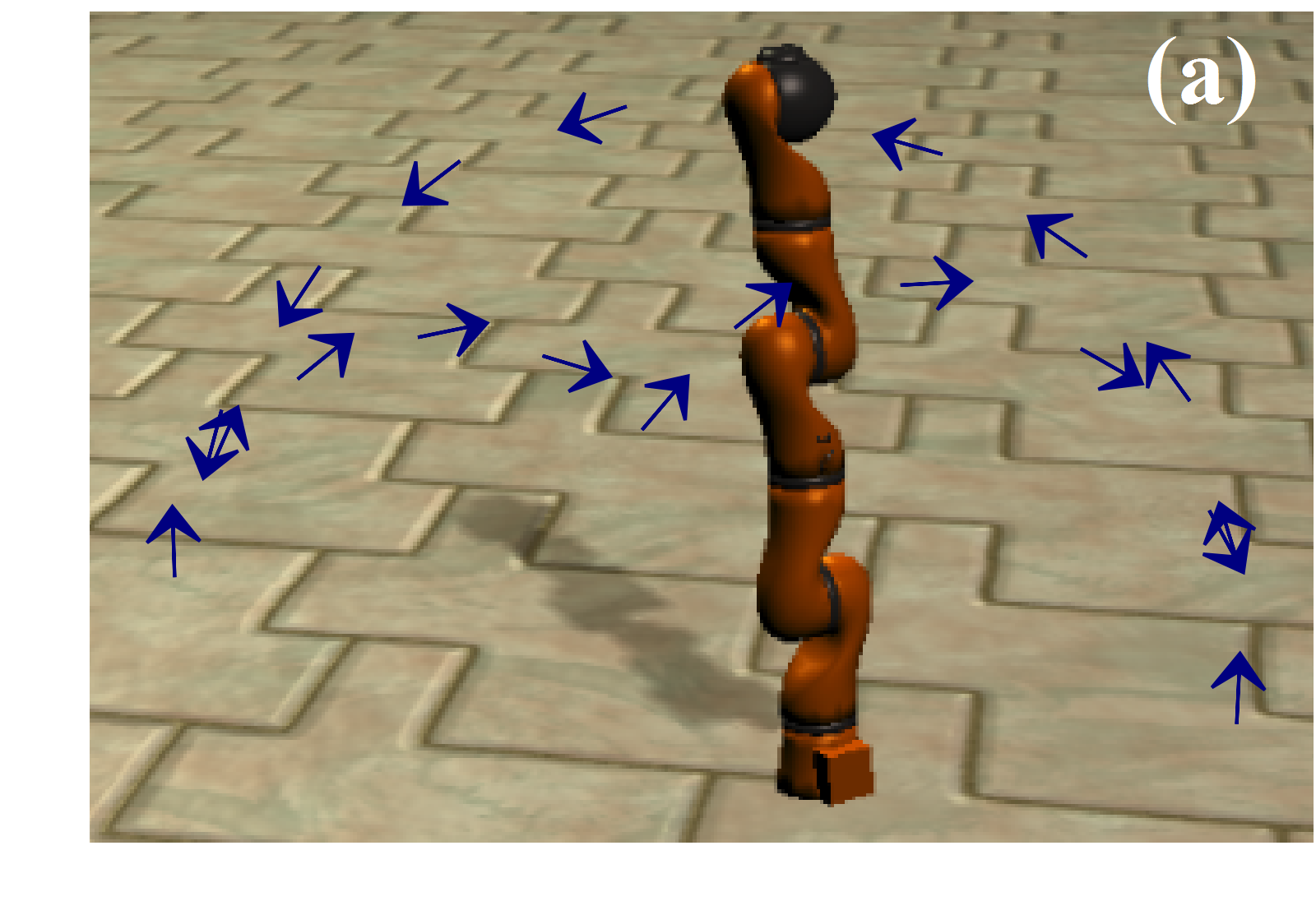}
  \includegraphics[width=0.5\linewidth]{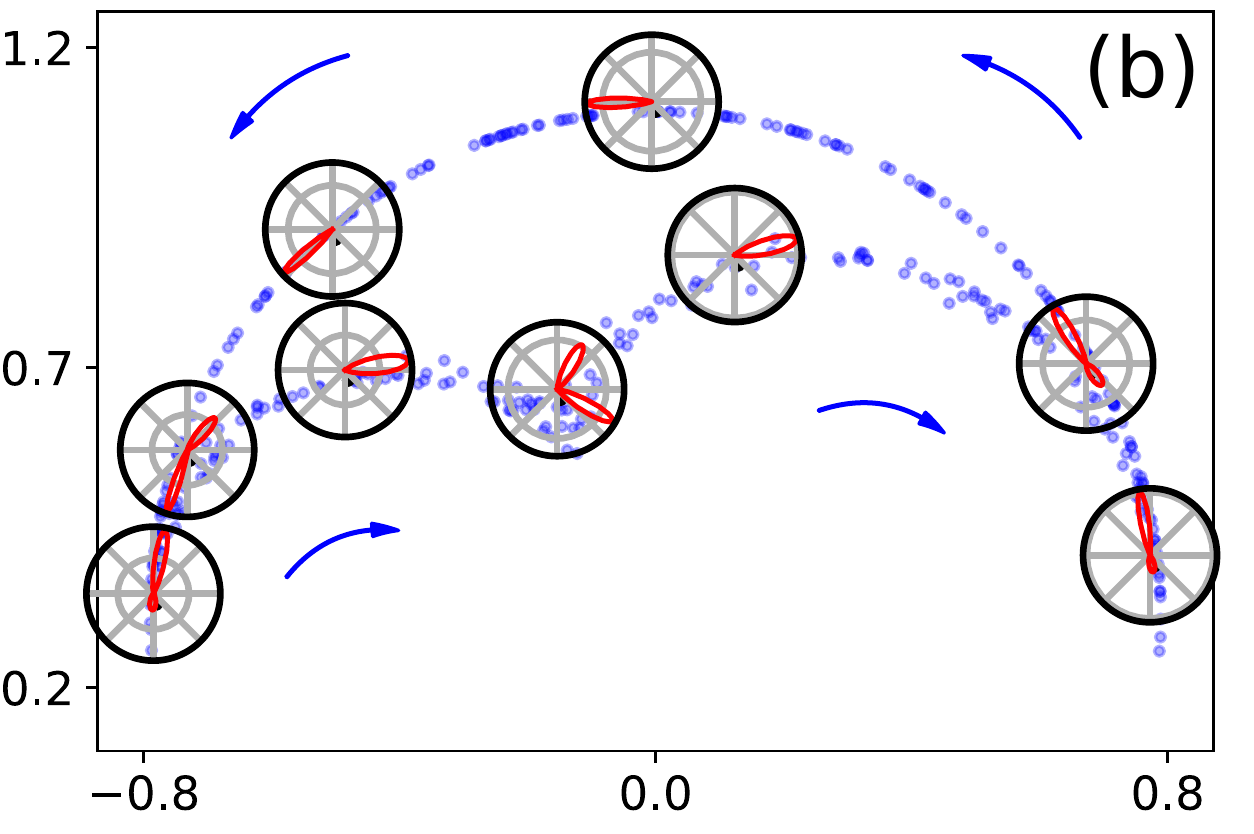}
  \caption{The Kuka robot arm (a) The approximate path of the end-effector is shown in blue arrows. Directional map of the Kuka end-effector movements. (b) Some of the data points collected over 20 such cycles with perturbation are shown in blue dots. The corresponding directional distribution at arbitrary locations are indicated by ploar plots.}
  \label{fig:kuka_1}
\end{figure}

In this section, we used Unimodal, Multi-modal, Edinburgh, and Kuka datasets to build the long-term spatiotemporal directional grid map so as to answer the question "What are the directions of movements in different places of the environment when observed over a period?". The resolutions of the grid maps were kept constant for demonstration and visualization purposes ($5\times4$, $5\times4$, and $17\times23$ cells for Unimodal, Multimodal, and Edinburgh datasets). The maps for Unimodal and Multimodal datasets are shown in Fig.~\ref{fig:moti} (b) and~\ref{fig:multi} (b), respectively. Around 15\% of randomly selected trajectories are shown in Fig.~\ref{fig:edin} (a) and the corresponding DGM in Fig.~\ref{fig:edin} (b). By comparing the trajectories and directions of lobes with intensities, it is possible to see the model has successfully learned the directions, including multi-modality. In order to demonstrate that the proposed algorithm is well suitable for other domains, we used the Kuka dataset. Additionally, as shown in Fig.~\ref{fig:kuka_1} (b), rather than maintaining a fixed grid, observations were taken from a few user-specified locations in the space.

\begin{figure}[h]
  \centering
  \includegraphics[width=0.7\linewidth]{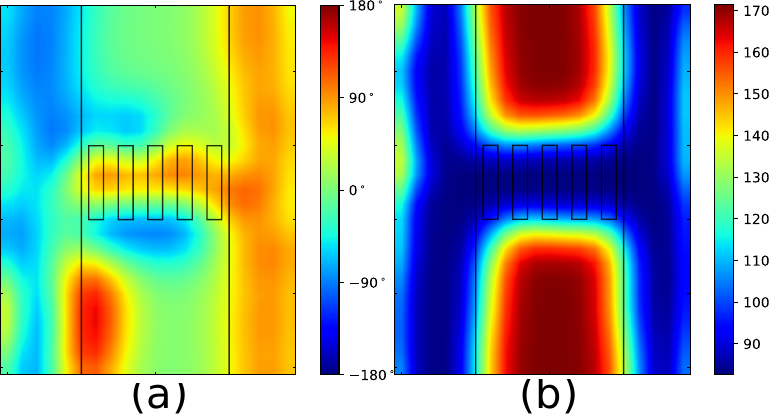}
  \caption{Gaussian process (GP) mapping \cite{human_simon} for Multimodal dataset where movements occur in different directions. This can be compared with Fig.~\ref{fig:multi}. (a) mean direction (b) confidence as variance. Observe that the predictions around places where multi-directional movements occur i.e. crosswalk and lower-left sidewalk is not accurate. In such places the GP averages all directions. As shown in (b), despite the inaccurate predictions the confidence about the prediction in such areas is also very high. Although angles are limited to $[-180^\circ, 180^\circ]$ in the figure, they can be in the range $(-\infty^\circ,\infty^\circ)$ without satisfying the recurrence relationship $f(\theta)=f(\theta+360^\circ)$.}
  \label{fig:gp}
\end{figure}

The quantitative aspects of different methods are given in Table~\ref{table:comp}. DGM-VMM is faster because 1) its means are initialized from DBSCAN, and 2) it has the flexibility to adjust to any shape. For the Unimodal dataset, the Gaussian process-based method proposed in \cite{human_simon} works well. As illustrated in Fig~\ref{fig:gp}, because the Gaussian process cannot handle multi-directional data, it merely averages directions, resulting in incorrect predictions. 

\begin{table*}[t]
  \centering
  \caption{Performance metrics for different methods. The smaller the ENLL or the higher the APD, the better the model is.}
    \begin{tabular}{p{0.09\linewidth}||c|c|c||c|c|c||c|c|c||c|c|c}
    \toprule
     &\multicolumn{3}{c||}{Unimodal dataset} &\multicolumn{3}{c||}{Multimodal dataset} &\multicolumn{3}{c||}{Edinburgh dataset} &\multicolumn{3}{c}{Kuka dataset} \\
       Method & ENLL & APD & Time & ENLL & APD & Time & ENLL & APD & Time & ENLL & APD & Time\\
      \hline
      DGM-VMM & {\bf 0.113} & 1.358 & {\bf15$\pm$13} & {\bf0.251} & {\bf1.015} & \bf{33$\pm$27} & {\bf1.483} & {\bf0.251} & {\bf28$\pm$25} & {\bf-0.087} & {\bf1.190} & {\bf13$\pm$9}\\
      DGM-VM & 0.177 & 1.172 & 71$\pm$34 & 0.696 & 0.615 & 115$\pm$62 & 1.733 & 0.202 & 72$\pm$72 & 0.747 & 0.699 & 55$\pm$15 \\
      GP \cite{human_simon}   & N/A    & {\bf1.711} & 278$\pm$16  & N/A   & 0.211 & 213$\pm$120 & N/A   & 0.124 & $>$3600 & N/A & 0.217 & 413$\pm$40 \\
      \bottomrule
    \end{tabular}
  \label{table:comp}
\end{table*}

\subsection{Experiment 3: Analyzing spatial variations}
This is merely a demonstration to show that the same model can be used to answer ``at a given time, where do everyone in the environment move?" For this purpose, all data points at a given time (i.e. $t=$fixed and all cells) in the Edinburgh dataset were considered and the EM algorithm was run for the von Mises mixture. The resulting polar plot is shown in Fig.~\ref{fig:edin} (c). To interpret, considering the entire environment, many people move towards $\approx-90^\circ$ and a few people $\approx90^\circ$. This kind of an analysis provides a summary statistic about the environment at a given time. Further, it is also possible to answer questions such as how fast the distribution changes over time by quantifying using mutual information or Kullback–Leibler divergence.

\subsection{Experiment 4: Analyzing temporal variations}
\begin{figure}[h]
    \centering
       \begin{subfigure}{\linewidth}
    \centering
    \includegraphics[width=0.9\linewidth]{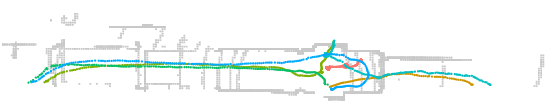}
    \caption{A robot and five humans move in a corridor simultaneously. The point cloud is shown in gray. \cite{romero2017inlida}}
    \end{subfigure}
    \begin{subfigure}{\linewidth}
    \centering
    \includegraphics[width=\linewidth]{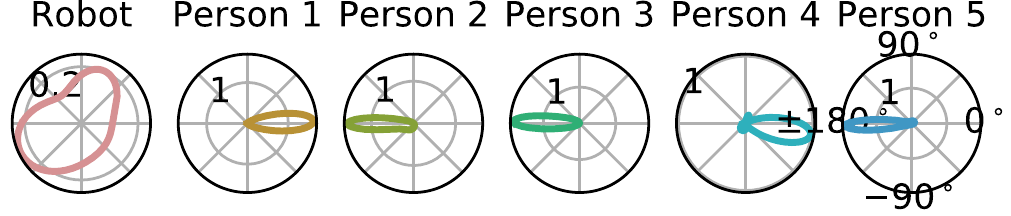}
    \caption{Directional distribution of each track is indicated by polar plots with corresponding colors.}
    \end{subfigure}
        \caption{Corridor dataset}
    \label{fig:s3}
    \vspace*{-10pt}
\end{figure}

\begin{figure}[h]
    \centering
    \begin{subfigure}{\linewidth}
    \centering
    \includegraphics[width=0.9\linewidth]{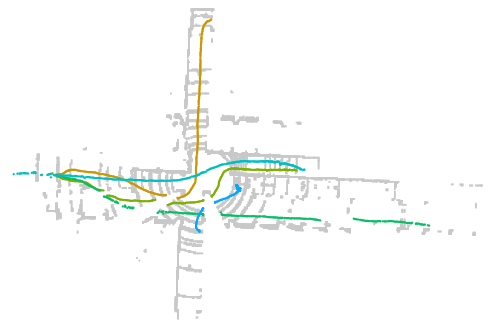}
    \caption{Five humans move in an intersection simultaneously. The point cloud is shown in gray. \cite{romero2017inlida}}
    \end{subfigure}
    \begin{subfigure}{\linewidth}
     \centering
    \includegraphics[width=0.8\linewidth]{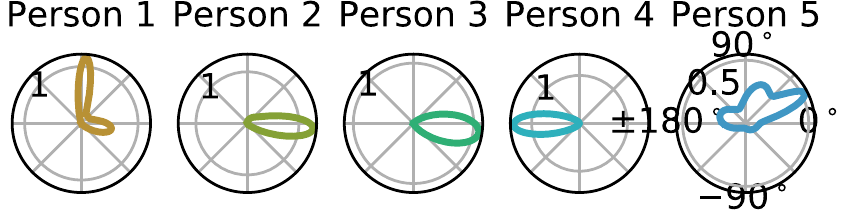}
    \caption{Directional distribution of each track is indicated by polar plots with corresponding colors.}
    \end{subfigure}
    \caption{Intersection dataset.}
    \label{fig:s5}
    \vspace*{-10pt}
\end{figure}

In this experiment, we analyze the temporal evolution of dynamic objects individually. For this purpose, we used the Corridor and Intersection datasets, and corresponding directional distributions are found in Figs.~\ref{fig:s3} and \ref{fig:s5}, respectively.


\begin{figure}[h]
  \centering
      \begin{subfigure}{0.49\linewidth}
     \centering
  \includegraphics[width=\linewidth]{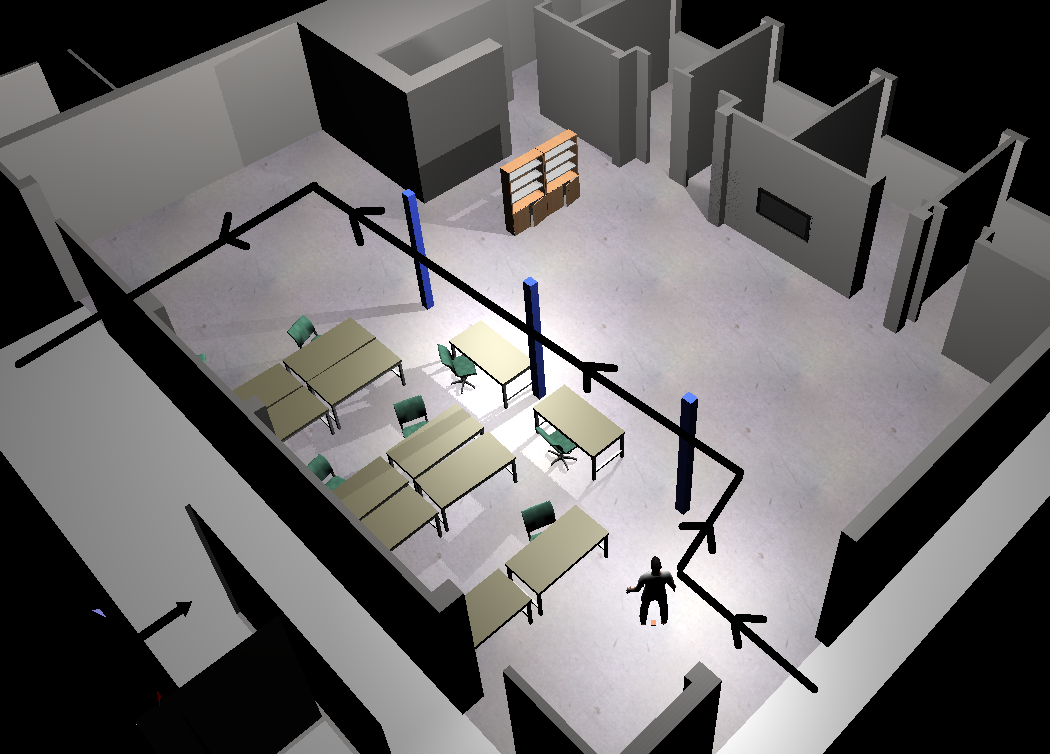}
  \caption{A person walking in the office environment.}
  \end{subfigure}
      \begin{subfigure}{0.49\linewidth}
     \centering
  \includegraphics[width=\linewidth]{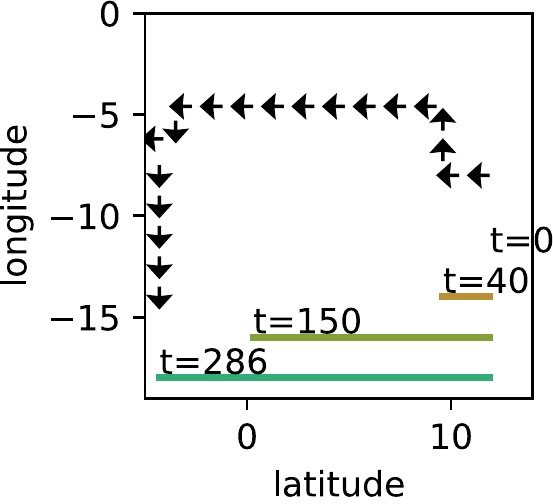}
  \caption{The plain view of the trajectory.}
  \end{subfigure}
   \begin{subfigure}{\linewidth}
     \centering
  \includegraphics[width=0.8\linewidth]{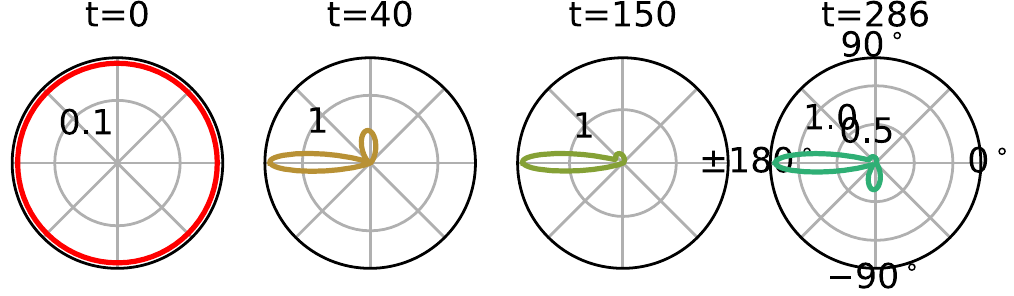}
  \caption{The directional distribution for four tome steps. Initially, it is a uniform circular distribution. Observe how the direction of lobes changes with the direction of trajectories at different time steps.}
  \end{subfigure}
  \caption{Human dataset}
  \label{fig:seql}
\end{figure}

Then, using the Human dataset, the temporal evolution of the map was analyzed in a sequential learning setting. In the office environment shown in Fig~\ref{fig:seql} (a), the trajectory of a simulated human is shown in Fig~\ref{fig:seql} (b). The model is learned in an online fashion as data are collected over 286 times steps. In Fig~\ref{fig:seql} (c), the directional distribution of four such time steps are shown. Starting with a dispersed distribution (i.e. any angle is possible or $\kappa \approx 0$), the observing robot sequentially learns the directions the human moves. 

\section{DISCUSSIONS}
\label{sec:dis}

\begin{figure}[h]
  \centering
    \includegraphics[width=0.49\linewidth]{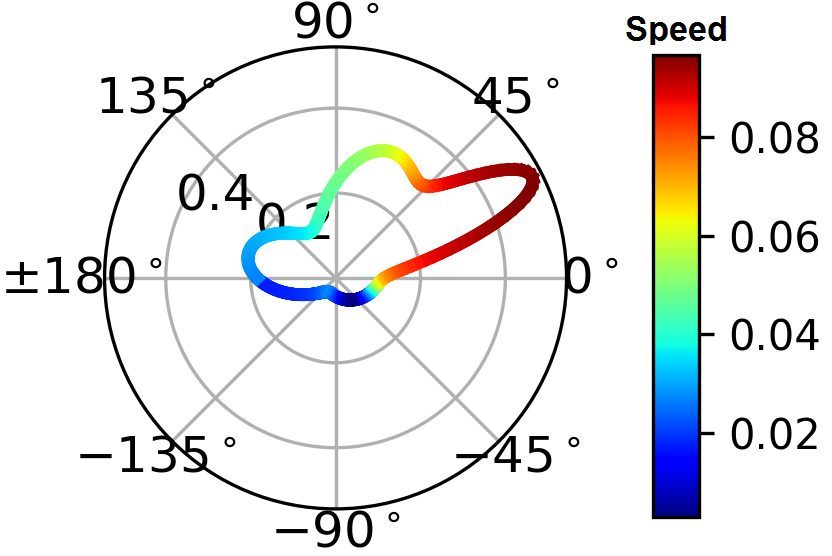}
    \includegraphics[width=0.49\linewidth]{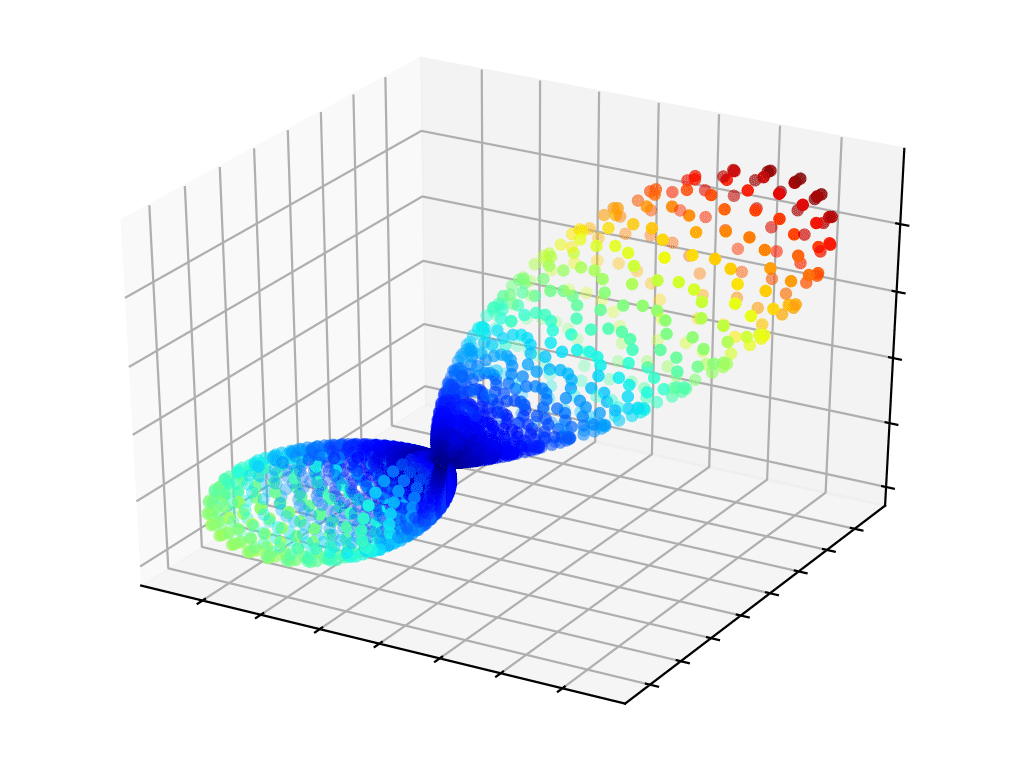}
  \caption{Usefulness of DGM (a) The directional distribution of Person 5 in Fig.~\ref{fig:s5} with colors indicating different speeds i.e. speed for different angles with the probability of angle (b) Demonstration of 3D lobes modeled using the von Mises-Fisher distribution.}
  \label{fig:diss}
\end{figure}

Similar to a Gaussian distribution, the mean, mode, and median of a unimodal von Mises is the same. However, when a mixture of von Mises is considered these quantities can be different. Especially, for practical applications, it is important to determine the modes. This is not straightforward because the values of $\kappa$ and $\mu$ determine both the number of modes and where they are. However, because von Mises belongs to the exponential family of distributions, it is possible to utilize similar algorithms that are used to find modes in the mixture of Gaussians \cite{carreira2000mode} or wavelet-based methods used in signal processing to find peaks \cite{du2006improved}.

In the proposed algorithm, we used DBSCAN to determine the number of mixture components. As this is not part of the EM algorithm, the number of mixture components maybe suboptimal. In order to further improve the likelihood, especially in high dimensional settings or when there is a small amount of data, taking a Bayesian approach, it is possible to consider $M$ as a parameter to be learned. Taking further advantages of the exponential family of distributions, as with the mixture of Gaussians \cite{Bishop2006}, it can be easily factorized and apply variational inference to jointly learn the number of mixtures as well as mixture parameters.

The one dimensional formulation can be easily extended to higher dimensions using the von Mises-Fisher extension given by \ref{eq:vonMF} for $(D-1)$ dimensions,
\begin{equation}
  \mathcal{VM}_D(\theta; \mu, \kappa) :=  \frac{\kappa^{D/2-1}}{ (2\pi)^{D/2} J_{D/2-1}(\kappa)} \exp{\big( \kappa \mu^\top \theta \big) }.
 \label{eq:vonMF}
\end{equation}
Such an extension can be used in modeling 3D spatiotemporal dynamics or modeling joint rotational uncertainties in robot manipulators with high degrees of freedom. By modeling bi-directional 3D uncertainties of a toy dataset (e.g. a moving drone in 3D), Fig.~\ref{fig:diss} (b) illustrates that such an extension is straightforward.

Interestingly, it is possible to use these directional maps along with other information extracted from dynamic environments to build maps that are informative. For instance, Fig.~\ref{fig:diss} (a) shows speeds (separately modeled) of different directions of "Person 5" in Fig.~\ref{fig:s5}. Additionally, a large DGM can be easily bundled together with an occupancy grid map to represent the dynamics of the environment well. These information-rich maps can then be used to make planning algorithms robust \cite{marinhofunctional,norouzi2016probabilistic}. 

One of the main limitations of the proposed method, as with any other grid-based method \cite{Elfes87}, the independence assumption among cells \cite{GPOMIJRR}. Therefore, it might be useful to consider the directional distribution as a conditional distribution or to have a more continuous representation \cite{mccalman2013multi,senanayake2017bayesian}.

\section{CONCLUSIONS}
\label{sec:conc}
We presented a robust algorithm to estimate angular uncertainties ubiquitous in robotics. To this end, we effectively we made use of directional statistics that are not typically utilized in robotics. Our method is generic enough to be used in any robotic platform such as mobile robots, drones, manipulators, etc. or in a variety of domains such as indoor mapping, field robotics, and human-robot interaction.








\bibliographystyle{IEEEtran.bst}
\bibliography{biblio}

\addtolength{\textheight}{-12cm}   


\end{document}